\documentclass[10pt]{cai26}

\usepackage{tabularx,multirow,array}
\usepackage[htt]{hyphenat}
\usepackage{seqsplit}
\usepackage{graphicx}
\usepackage{fancybox}
\usepackage[table,dvipsnames]{xcolor}
\usepackage{amssymb}
\usepackage{pifont}
\usepackage{booktabs,array,adjustbox}
\usepackage{colortbl}
\usepackage{tikz}
\usepackage{array}
\usepackage{makecell}
\usepackage{geometry}

\makeatletter
\newcommand{\supertiny}{%
  \ifcase\@ptsize
    \fontsize{3}{4}\selectfont 
  \or
    \fontsize{4}{5}\selectfont 
  \or
    \fontsize{4}{5}\selectfont 
  \fi
}
\makeatother

\begin{document}
\def\conferenceyear{2026}
\volumeheader{39}{0}
\begin{center}

\title{Multi-Dimensional Model Integrity and Responsibility Assessment Index and Scoring Framework}

\maketitle

\thispagestyle{empty}
\pagenumbering{gobble}

\begin{tabular}{cc}
Phuc Truong Loc Nguyen\upstairs{\affilone*}, Thanh Hung Do\upstairs{\affilone}, \\ Truong Thanh Hung Nguyen\upstairs{\affiltwo}, Hung Cao\upstairs{\affiltwo}
\\[0.25ex]
{\small \upstairs{\affilone} Friedrich-Alexander-Universität Erlangen-Nürnberg, Germany} \\
{\small \upstairs{\affiltwo} University of New Brunswick, Canada} \\
\end{tabular}

\emails{
  \upstairs{*}Corresponding author: loc.pt.nguyen@fau.de 
}
\vspace*{0.1in}
\end{center}

\begin{abstract}
Artificial intelligence in high-stakes tabular domains cannot be evaluated by predictive performance alone, yet current practice still assesses explainability, fairness, robustness, privacy, and sustainability mostly in isolation. We propose the Model Integrity and Responsibility Assessment Index (MIRAI), a unified evaluation framework that measures tabular models across these five dimensions under a controlled comparison setting and aggregates them into a single score. MIRAI combines established metrics through normalized and direction-aligned dimension scores, which enables direct comparison across models with different architectural and computational profiles. Experiments on healthcare, financial, and socioeconomic datasets show that higher predictive performance does not necessarily imply better overall integrity and responsibility. In several cases, simpler models achieve a stronger cross-dimensional balance than more complex deep tabular architectures. MIRAI provides a compact and practical basis for responsible model selection in regulated settings.
\end{abstract}

\copyrightnotice

\section{Introduction}

Artificial Intelligence (AI) is increasingly used in high-stakes domains such as healthcare, finance, and public decision support, where model outputs can directly affect individuals and institutions \cite{rbbb,rccc,rhhh,raaa,rggg,rrrr}. In these settings, tabular data remains a dominant format for predictive modeling \cite{r1,r21}. Predictive performance alone is not sufficient for deployment, because a highly accurate model may still be difficult to interpret, unfair across demographic groups, vulnerable to perturbations, prone to privacy leakage, or expensive to operate. These properties often conflict, which makes responsible model selection inherently multi-dimensional \cite{r5,r19,r1918}. Yet current evaluation practice remains fragmented. Explainability, fairness, robustness, privacy, and sustainability are still assessed largely in isolation, often with separate metrics and toolchains \cite{r7,r13,r14}. Recent integrated efforts move toward broader responsible-AI assessment \cite{r5,r18,r6}, but model selection still lacks a compact framework that can translate heterogeneous evidence into a clear comparative judgment \cite{r4,r19,r27}.

We address this gap with the \textit{Model Integrity and Responsibility Assessment Index (MIRAI)}, a unified evaluation framework for tabular models. MIRAI quantifies explainability, fairness, sustainability, robustness, and privacy, aligns heterogeneous metrics to a common scoring direction, and aggregates them into a single comparative index. This provides a compact and decision-oriented basis for responsible model assessment in regulated settings \cite{r4,r5}. Experiments on healthcare, financial, and socioeconomic datasets show that stronger predictive performance does not necessarily imply better overall integrity and responsibility. In several cases, simpler models achieve a better cross-dimensional balance than more complex deep tabular architectures.
\section{Background and Related Work}

Recent work has established a rich set of metrics and toolkits for evaluating responsible AI properties at the dimension level. In explainability, quantitative evaluation has moved beyond visual inspection toward protocol-based assessment of faithfulness, robustness, randomization, and complexity, with Quantus becoming a widely used benchmark framework \cite{r7,r9}. Fairness evaluation is similarly supported by mature libraries such as Fairlearn and AIF360, while robustness and privacy are commonly examined through adversarial testing and leakage analysis frameworks such as ART and related toolkits \cite{r13,r14,r15}. Sustainability is mostly assessed through carbon and computational cost accounting, which extends evaluation beyond predictive utility to environmental and resource efficiency \cite{r12,r20}.

Building on these foundations, several studies have moved toward integrated responsible-AI assessment. RAISE proposes a unified scoring pipeline for tabular models across multiple dimensions \cite{r5}, while SAFE introduces an internally consistent integrated metric formulation \cite{r6}. Complementary compliance-oriented frameworks further stress the need for reproducible and auditable assessment pipelines in regulated settings such as those targeted by the EU AI Act \cite{r18,r27}. Yet an important gap remains. Existing approaches still vary in dimension coverage, metric aggregation, and comparison protocols, while trade-offs across fairness, explainability, privacy, robustness, and efficiency can substantially alter model rankings \cite{r5,r19}. This makes deployment judgments difficult when raw metrics point in different directions. MIRAI is designed to address this gap through a compact, controlled, and directly comparative evaluation framework for tabular models.
\section{MIRAI Scoring Framework}\label{frame}

The MIRAI framework performs controlled comparative evaluation for tabular binary classification. It requires at least two candidate models from major tabular learning families, including Decision Tree (DT), XGBoost (XGB), Support Vector Machine (SVM), Multilayer Perceptron (MLP), TabResNet (TRN), and Feature Tokenizer Transformer (FTT) \cite{gorishniy2021revisiting}. For each candidate, it supports model-specific architectural settings, such as tree depth and hidden dimensions, together with training hyperparameters. It also records key dataset properties, including sample size, feature dimensionality, domain, and task type. In addition, it designates a Target Model as the intended deployment candidate and uses it as the reference for ranking and comparison. This makes the evaluation decision-oriented, since model quality, computational overhead, and cross-dimensional trade-offs can be judged relative to a concrete baseline and used to support model and hyperparameter refinement.

Each model is evaluated across five dimensions: explainability, fairness, sustainability, robustness, and privacy. Explainability is computed in two stages. SHAP \cite{NIPS2017_8a20a862} first generates feature attributions, and Quantus \cite{r7} then evaluates them with eight metrics: Local Lipschitz Estimate \cite{alvarez2018robustness} and Consistency \cite{pmlr-v162-dasgupta22a} for robustness, Faithfulness Correlation \cite{ijcai2020p417} and Faithfulness Estimate \cite{NEURIPS2018_3e9f0fc9} for faithfulness, Model Parameter Randomization Test \cite{NEURIPS2018_294a8ed2} and Random Logit Test \cite{pmlr-v119-sixt20a} for randomization, and Sparseness \cite{pmlr-v119-chalasani20a} and Complexity \cite{ijcai2020p417} for complexity. Fairness is assessed through subgroup performance in Accuracy, Precision, Recall, True Positive Rate, and False Positive Rate, followed by absolute inter-group disparities, Demographic Parity \cite{doi:10.1177/0049124118782533}, and Equalized Odds \cite{NIPS2016_6a9659fe}, using Fairlearn \cite{r13} and AIF360 \cite{r14}. Sustainability combines carbon emissions, parameter count, FLOPs, and MACs. Carbon impact is estimated with the Lacoste score \cite{r12}, where \(p_c\) is average CPU power, \(p_g\) is average GPU power, and \(p_t=(p_c+p_g)/1000\) is total power in kilowatts. This value is multiplied by the 2023 Canadian electricity emission rate \cite{eccc_annex13_electricity_intensity_2025} and then normalized by the 2023 daily per-capita emission reference \cite{eccc_cesi_ghg_emissions_2025}. Robustness is measured through the HopSkipJump Attack \cite{chen2019boundary} accuracy gap with ART \cite{r15} and prediction-space Maximum Mean Discrepancy drift detection with Alibi-Detect \cite{alibi-detect}. Privacy is quantified through Membership Inference Privacy \cite{7958568} and SHAPr Privacy \cite{duddu2021shapr}, both implemented with ART.

All raw metrics are normalized to \([0,1]\), where 1 denotes the most desirable outcome. Metrics for which lower raw values are preferable are direction-aligned through \(1-\mathrm{raw}\). The normalized metrics within each dimension are averaged to obtain a Dimension Score, and the final MIRAI score is computed as \( \mathrm{MIRAI}=\sum_{d=1}^{5} w_d \mathrm{DS}_d \). Equal weights are used by default, with \(w_d=0.2\), but user-defined weights are also supported to reflect application-specific priorities, such as fairness in regulated settings or robustness in safety-critical deployment. Predictive Accuracy and F1-score are reported separately, so MIRAI complements rather than replaces standard performance measures. This formulation preserves dimension-specific evidence while enabling compact, context-aware, and directly comparative model selection.

\section{Experiment and Results} \label{exp_res}

We evaluate MIRAI on six classifiers that span major tabular learning paradigms: DT, XGB, SVM, MLP, TRN, and FTT. The evaluation uses three public high-stakes tabular datasets from healthcare, finance, and socioeconomics: Diabetes Hospitals \cite{diabetes_130-us_hospitals_for_years_1999-2008_296}, German Credit \cite{statlog}, and Census Income \cite{adult_2}. All models are trained under controlled conditions to ensure fair comparison. For fairness evaluation, ``gender'' is used as the sensitive attribute and ``male'' is treated as the privileged group, following the protocol in \cite{r14,10.1257/app.20210180}. MIRAI then compares the models jointly across explainability, fairness, sustainability, robustness, and privacy. Results are presented in Table~\ref{tab:diabetes_results}, Table~\ref{tab:german_results}, and Table~\ref{tab:census_results}.

Across datasets, MIRAI reveals a consistent trade-off between predictive strength and cross-dimensional model quality. Deep tabular models, especially TRN and FTT, remain competitive in predictive performance and can achieve strong explainability, with TRN reaching top-tier explainability on Census Income and German Credit. However, these gains are often offset by weaker sustainability and privacy, especially for FTT, whose high computational cost leads to severe efficiency penalties. By contrast, MLP and XGB maintain very strong sustainability, while SVM shows the strongest privacy behavior on Diabetes and German Credit. Fairness and robustness also depend on the data regime. On Diabetes, most models achieve both high robustness and high fairness, with XGB reaching near-perfect fairness. On smaller or more imbalanced settings, such as German Credit, deeper models lose ground in fairness and privacy.

The key result is that predictive metrics and MIRAI rankings do not coincide. Models with the best F1 scores do not necessarily provide the strongest overall integrity and responsibility profile when all five dimensions are considered jointly. On German Credit, TRN attains the top F1 score, yet MLP achieves the stronger MIRAI ranking. On Diabetes, MLP records the highest MIRAI score, while FTT ranks much lower despite competitive predictive performance. Across the benchmarks, MLP and SVM provide the most favorable cross-dimensional balance. This indicates that higher model complexity does not guarantee better all-round behavior and that simpler models can be stronger candidates for deployment in regulated tabular settings.


\begin{table}[ht]
\centering
\caption{\textbf{Diabetes Hospitals}: 101763 samples, 22 features. The best results are in \textbf{bold}. The second-best results are \underline{underlined}.}
\label{tab:diabetes_results}
\supertiny
\setlength{\tabcolsep}{1pt}
\renewcommand{\arraystretch}{1}

\definecolor{themeblue}{RGB}{20, 85, 215}   
\definecolor{themelight}{RGB}{235, 240, 250}  
\arrayrulecolor{themeblue} 

\begin{tabularx}{\linewidth}{>{\raggedright\arraybackslash}p{3.6cm}*{6}{>{\centering\arraybackslash}X}}

\textbf{Model} & \textbf{DT} & \textbf{XGB} & \textbf{SVM} & \textbf{MLP} & \textbf{TRN} & \textbf{FTT} \\
\hline
\hline

\rowcolor{themelight}
\cellcolor{themeblue}\textcolor{white}{\textbf{MIRAI}}
& 0.7635 & \underline{0.7763} & 0.7724 & \textbf{0.7776} & 0.7607 & 0.5636 \\
Accuracy
& 0.7870 & 0.8880 & 0.8880 & 0.8840 & 0.8850 & 0.8880 \\
F1 Score
& 0.7960 & 0.8360 & 0.8360 & 0.8380 & 0.8370 & 0.8360 \\
\hline

\rowcolor{themelight}\textcolor{themeblue}{\textbf{Explainability}}
& 0.4456 & \textbf{0.5126} & 0.4312 & 0.4850 & \underline{0.5101} & 0.4635 \\
Complexity
& 0.5619 & 0.6363 & 0.6237 & 0.6454 & 0.6988 & 0.6639 \\
Faithfulness
& 0.6105 & 0.7211 & 0.5285 & 0.5318 & 0.5571 & 0.6910 \\
Robustness (Expl.)
& 0.1101 & 0.1931 & 0.0725 & 0.2016 & 0.1730 & 0.1260 \\
Randomization
& 0.5000 & 0.5000 & 0.5000 & 0.5611 & 0.6117 & 0.3732 \\
\hline

\rowcolor{themelight}\textcolor{themeblue}{\textbf{Fairness}}
& \underline{0.9980} & \textbf{0.9993} & 0.9645 & 0.9947 & 0.9887 & 0.9155 \\
Accuracy Diff*
& 0.0020 & 0.0040 & 0.0050 & 0.0040 & 0.0030 & 0.0040 \\
Precision Diff*
& 0.0040 & 0.0000 & 0.2000 & 0.0240 & 0.0410 & 0.5000 \\
TPR Diff*
& 0.0020 & 0.0000 & 0.0030 & 0.0020 & 0.0110 & 0.0010 \\
FPR Diff*
& 0.0010 & 0.0000 & 0.0010 & 0.0000 & 0.0000 & 0.0010 \\
Demographic Parity Diff*
& 0.0010 & 0.0000 & 0.0010 & 0.0000 & 0.0020 & 0.0000 \\
Equalized Odds Diff*
& 0.0020 & 0.0000 & 0.0030 & 0.0020 & 0.0110 & 0.0010 \\
\hline

\rowcolor{themelight}\textcolor{themeblue}{\textbf{Sustainability}}
& 0.9899 & \textbf{0.9992} & 0.9639 & \underline{0.9987} & 0.8913 & 0.0000 \\
Parameter Count*
& 0.9698 & 0.9976 & 0.8988 & 0.9964 & 0.7022 & 0.0000 \\
FLOPs per Sample*
& 1.0000 & 1.0000 & 0.9976 & 0.9998 & 0.9861 & 0.0000 \\
MACs per Sample*
& 1.0000 & 1.0000 & 0.9952 & 0.9998 & 0.9857 & 0.0000 \\
Normalized kgCO$_2$e*
& 0.9925 & 0.9989 & 0.9720 & 0.9987 & 0.8850 & 0.0000 \\
\hline

\rowcolor{themelight}\textcolor{themeblue}{\textbf{Robustness}}
& \underline{0.8676} & 0.8619 & 0.8665 & 0.8506 & 0.8553 & \textbf{0.8730} \\
HSJA Robustness
& 0.9558 & 0.9257 & 0.9304 & 0.8889 & 0.9058 & 0.9237 \\
Drift Robustness
& 0.7794 & 0.7980 & 0.8027 & 0.8123 & 0.8048 & 0.8224 \\
\hline

\rowcolor{themelight}\textcolor{themeblue}{\textbf{Privacy}}
& 0.5164 & 0.5144 & \textbf{0.6361} & 0.5590 & 0.5582 & \underline{0.5635}\\
MI Privacy
& 0.4762 & 0.4813 & 0.7158 & 0.5666 & 0.5639 & 0.5724 \\
SHAPr Privacy
& 0.5566 & 0.5475 & 0.5565 & 0.5514 & 0.5526 & 0.5545 \\
\hline

\multicolumn{7}{@{}l@{}}{\scriptsize\parbox[t]{\linewidth}{Metrics marked with an asterisk (*) are lower-is-better by definition. Their values have been inverted using $1-\text{raw}$.}} 
\end{tabularx}
\arrayrulecolor{black} 
\end{table}

\begin{table}[ht]
\centering
\caption{\textbf{German Credit}: 1000 samples, 22 features. The best results are in \textbf{bold}. The second-best results are \underline{underlined}.}
\label{tab:german_results}

\supertiny
\setlength{\tabcolsep}{1pt}
\renewcommand{\arraystretch}{1.3}

\definecolor{themeblue}{RGB}{20, 85, 215}   
\definecolor{themelight}{RGB}{235, 240, 250}  
\arrayrulecolor{themeblue} 

\begin{tabularx}{\linewidth}{>{\raggedright\arraybackslash}p{3.6cm}*{6}{>{\centering\arraybackslash}X}}
\textbf{Model} & \textbf{DT} & \textbf{XGB} & \textbf{SVM} & \textbf{MLP} & \textbf{TRN} & \textbf{FTT} \\
\hline
\hline
\rowcolor{themelight}
\cellcolor{themeblue}\textcolor{white}{\textbf{MIRAI}}
& 0.7282 & 0.7086 & \underline{0.7377} & \textbf{0.7422} & 0.6540 & 0.4815 \\
Accuracy
& 0.7000 & 0.7550 & 0.7100 & 0.7350 & 0.7500 & 0.7350 \\
F1 Score
& 0.7070 & 0.7460 & 0.6910 & 0.7330 & 0.7520 & 0.7340 \\
\hline

\rowcolor{themelight}\textcolor{themeblue}{\textbf{Explainability}}
& 0.4601 & 0.4371 & 0.4451 & \underline{0.4951} & \textbf{0.4982} & 0.4690 \\
Complexity
& 0.6065 & 0.5606 & 0.6072 & 0.6154 & 0.6104 & 0.6343 \\
Faithfulness
& 0.5265 & 0.4981 & 0.5395 & 0.5585 & 0.4722 & 0.5045 \\
Robustness (Expl.)
& 0.2074 & 0.1898 & 0.1339 & 0.1797 & 0.1651 & 0.1656 \\
Randomization
& 0.5000 & 0.5000 & 0.5000 & 0.6270 & 0.7451 & 0.5716 \\
\hline

\rowcolor{themelight}
\textcolor{themeblue}{\textbf{Fairness}}
& 0.8907 & \textbf{0.9465} & 0.9017 & 0.8862 & 0.8170 & \underline{0.9202} \\
Accuracy Diff*
& 0.1210 & 0.0550 & 0.0450 & 0.0190 & 0.1520 & 0.0620 \\
Precision Diff*
& 0.0790 & 0.0420 & 0.0190 & 0.0160 & 0.0580 & 0.0370 \\
TPR Diff*
& 0.1650 & 0.0650 & 0.1000 & 0.1000 & 0.2650 & 0.1000 \\
FPR Diff*
& 0.0000 & 0.0250 & 0.1500 & 0.2000 & 0.1250 & 0.0750 \\
Demographic Parity Diff*
& 0.1260 & 0.0690 & 0.1260 & 0.1480 & 0.2330 & 0.1050 \\
Equalized Odds Diff*
& 0.1650 & 0.0650 & 0.1500 & 0.2000 & 0.2650 & 0.1000 \\
\hline

\rowcolor{themelight}
\textcolor{themeblue}{\textbf{Sustainability}}
& \textbf{0.9999} & \underline{0.9993} & 0.9951 & 0.9987 & 0.8913 & 0.0000 \\
Parameter Count*
& 0.9996 & 0.9978 & 0.9863 & 0.9964 & 0.7022 & 0.0000 \\
FLOPs per Sample*
& 1.0000 & 1.0000 & 0.9997 & 0.9998 & 0.9861 & 0.0000 \\
MACs per Sample*
& 1.0000 & 1.0000 & 0.9993 & 0.9998 & 0.9857 & 0.0000 \\
Normalized kgCO$_2$e*
& 0.9992 & 0.9985 & 0.9920 & 0.9978 & 0.8900 & 0.0000 \\
\hline

\rowcolor{themelight}
\textcolor{themeblue}{\textbf{Robustness}}
& 0.6715 & 0.5308 & \underline{0.6830} & \textbf{0.6930} & 0.4927 & 0.4853 \\
HSJA Robustness
& 0.5900 & 0.2700 & 0.5750 & 0.5650 & 0.2300 & 0.2000 \\
Drift Robustness
& 0.7530 & 0.7916 & 0.7911 & 0.8210 & 0.7555 & 0.7706 \\
\hline

\rowcolor{themelight}
\textcolor{themeblue}{\textbf{Privacy}}
& 0.6188 & 0.6295 & \textbf{0.6635} & \underline{0.6382} & 0.5706 & 0.5329 \\
MI Privacy
& 0.5333 & 0.7095 & 0.7000 & 0.6476 & 0.5809 & 0.5476 \\
SHAPr Privacy
& 0.7042 & 0.5495 & 0.6270 & 0.6289 & 0.5602 & 0.5181 \\
\hline

\multicolumn{7}{@{}l@{}}{\scriptsize\parbox[t]{\linewidth}{Metrics marked with an asterisk (*) are lower-is-better by definition. Their values have been inverted using $1-\text{raw}$.}} 

\end{tabularx}
\arrayrulecolor{black}
\end{table}

\begin{table}[ht]
\centering
\caption{\textbf{Census Income}: 32561 samples, 14 features. The best results are in \textbf{bold}. The second-best results are \underline{underlined}.}
\label{tab:census_results}

\supertiny
\setlength{\tabcolsep}{1pt}
\renewcommand{\arraystretch}{1.3}

\definecolor{themeblue}{RGB}{20, 85, 215}
\definecolor{themelight}{RGB}{235, 240, 250}
\arrayrulecolor{themeblue}

\begin{tabularx}{\linewidth}{>{\raggedright\arraybackslash}p{3.6cm}*{6}{>{\centering\arraybackslash}X}}
\textbf{Model} & \textbf{DT} & \textbf{XGB} & \textbf{SVM} & \textbf{MLP} & \textbf{TRN} & \textbf{FTT} \\
\hline
\hline

\rowcolor{themelight}
\cellcolor{themeblue}\textcolor{white}{\textbf{MIRAI}}
& 0.6925 & 0.6890 & \textbf{0.7209} & \underline{0.7189} & 0.6881 & 0.5698 \\
Accuracy
& 0.8130 & 0.8690 & 0.8530 & 0.8520 & 0.8500 & 0.8490 \\
F1 Score
& 0.8140 & 0.8630 & 0.8440 & 0.8500 & 0.8460 & 0.8480 \\
\hline

\rowcolor{themelight}
\textcolor{themeblue}{\textbf{Explainability}}
& 0.4491 & 0.4271 & 0.4547 & \underline{0.5078} & \textbf{0.5250} & 0.4809 \\
Complexity
& 0.6287 & 0.6259 & 0.6334 & 0.6014 & 0.6704 & 0.7005 \\
Faithfulness
& 0.5339 & 0.4810 & 0.5856 & 0.5864 & 0.5963 & 0.6026 \\
Robustness (Expl.)
& 0.1339 & 0.1016 & 0.0999 & 0.1171 & 0.0975 & 0.1003 \\
Randomization
& 0.5000 & 0.5000 & 0.5000 & 0.7262 & 0.7358 & 0.5200 \\
\hline

\rowcolor{themelight}
\textcolor{themeblue}{\textbf{Fairness}}
& 0.9035 & 0.9012 & 0.9117 & 0.9168 & \underline{0.9210} & \textbf{0.9387} \\
Accuracy Diff*
& 0.1210 & 0.1010 & 0.1130 & 0.1150 & 0.1060 & 0.1100 \\
Precision Diff*
& 0.0860 & 0.0590 & 0.0520 & 0.0090 & 0.0640 & 0.0050 \\
TPR Diff*
& 0.0000 & 0.0990 & 0.0730 & 0.0490 & 0.0120 & 0.0170 \\
FPR Diff*
& 0.0970 & 0.0630 & 0.0620 & 0.0770 & 0.0660 & 0.0480 \\
Demographic Parity Diff*
& 0.1780 & 0.1720 & 0.1570 & 0.1720 & 0.1600 & 0.1400 \\
Equalized Odds Diff*
& 0.0970 & 0.0990 & 0.0730 & 0.0770 & 0.0660 & 0.0480 \\
\hline

\rowcolor{themelight}
\textcolor{themeblue}{\textbf{Sustainability}}
& 0.9971 & \textbf{0.9992} & 0.9779 & \underline{0.9991} & 0.8864 & 0.0000 \\
Parameter Count*
& 0.9914 & 0.9976 & 0.9403 & 0.9976 & 0.7023 & 0.0000 \\
FLOPs per Sample*
& 1.0000 & 1.0000 & 0.9978 & 0.9998 & 0.9787 & 0.0000 \\
MACs per Sample*
& 1.0000 & 1.0000 & 0.9957 & 0.9998 & 0.9781 & 0.0000 \\
Normalized kgCO$_2$e*
& 0.9968 & 0.9990 & 0.9735 & 0.9988 & 0.8840 & 0.0000 \\
\hline

\rowcolor{themelight}
\textcolor{themeblue}{\textbf{Robustness}}
& 0.5948 & 0.5092 & \underline{0.6406} & 0.5967 & 0.5160 & \textbf{0.8607} \\
HSJA Robustness
& 0.3740 & 0.3840 & 0.4100 & 0.3560 & 0.4300 & 0.9280 \\
Drift Robustness
& 0.8156 & 0.6345 & 0.8712 & 0.8373 & 0.6020 & 0.7934 \\
\hline

\rowcolor{themelight}
\textcolor{themeblue}{\textbf{Privacy}}
& 0.5171 & \underline{0.6098} & \textbf{0.6166} & 0.5710 & 0.5922 & 0.5700 \\
MI Privacy
& 0.4767 & 0.6541 & 0.6729 & 0.5813 & 0.6037 & 0.5841 \\
SHAPr Privacy
& 0.5574 & 0.5654 & 0.5602 & 0.5607 & 0.5807 & 0.5559 \\
\hline

\multicolumn{7}{@{}l@{}}{\scriptsize\parbox[t]{\linewidth}{Metrics marked with an asterisk (*) are lower-is-better by definition. Their values have been inverted using $1-\text{raw}$.}} 

\end{tabularx}
\arrayrulecolor{black}
\end{table}

\section{Conclusion}
We presented MIRAI, a unified evaluation framework for tabular models that integrates explainability, fairness, sustainability, robustness, and privacy beyond predictive performance alone. Our results show that higher model complexity does not necessarily translate into better overall integrity and responsibility. In several high-stakes settings, simpler models such as MLP and SVM achieve a stronger cross-dimensional balance than transformer-based architectures. These findings suggest that compact, well-balanced models can be better suited to deployment in regulated domains.

\clearpage
\section*{Acknowledgment}
This work was supported by NSERC Discovery Grant No RGPIN-2025-04478 and NSERC Discovery Supplement Award No DGECR-2025-00129.



\printbibliography[heading=subbibintoc]


\end{document}